\documentclass[conference]{IEEEtran}
\IEEEoverridecommandlockouts
\usepackage{cite}
\usepackage{amsmath,amssymb,amsfonts}
\usepackage{algorithmic}
\usepackage{graphicx}
\usepackage{textcomp}
\usepackage{xcolor}
\usepackage{booktabs}
\usepackage{url}
\usepackage[colorlinks=true,allcolors=blue]{hyperref}%
\usepackage{tablefootnote}
\usepackage{subcaption}
\usepackage{enumitem}
\usepackage{adjustbox}

\usepackage[linesnumbered,ruled,vlined]{algorithm2e}
\SetKwInput{kwInit}{Init}
\SetKwInput{kwEval}{Evaluate}
\SetKwInput{kwCreate}{Create}
\SetKwComment{Comment}{$\triangleright$\ }{}

\def\BibTeX{{\rm B\kern-.05em{\sc i\kern-.025em b}\kern-.08em
    T\kern-.1667em\lower.7ex\hbox{E}\kern-.125emX}}
\begin{document}

\title{Quality and Computation Time in Optimization Problems \\
}

\author{\IEEEauthorblockN{Zhicheng He}
\IEEEauthorblockA{\textit{Department of Computer Science and Engineering} \\
\textit{Southern University of Science and Technology}\\
Shenzhen, China \\
11612601@mail.sustech.edu.cn}
}

\maketitle

\begin{abstract}
Optimization problems are crucial in artificial intelligence. 
Optimization algorithms are generally used to adjust the performance of artificial intelligence models to minimize the error of mapping inputs to outputs.
Current evaluation methods on optimization algorithms generally consider the performance in terms of quality.
However, not all optimization algorithms for all test cases are evaluated equal from quality, the computation time should be also considered for optimization tasks.
In this paper, we investigate the quality and computation time of optimization algorithms in optimization problems, instead of the one-for-all evaluation of quality.
We select the well-known optimization algorithms (Bayesian optimization and evolutionary algorithms) and evaluate them on the benchmark test functions in terms of quality and computation time.
The results show that BO is suitable to be applied in the optimization tasks that are needed to obtain desired quality in the limited function evaluations, and the EAs are suitable to search the optimal of the tasks that are allowed to find the optimal solution with enough function evaluations.
This paper provides the recommendation to select suitable optimization algorithms for optimization problems with different numbers of function evaluations, which contributes to the efficiency that obtains the desired quality with less computation time for optimization problems.

\end{abstract}


\section{Introduction}

To date, optimization problems are crucial in many applications, particularly the artificial intelligence.
An optimization problem is the task of finding the desired solution from the feasible solution space.
Optimization algorithms have been widely to search the desired solutions for the optimization problems.
The current optimization algorithms are generally trial and error methods.
A trial and error method of problem solving where a candidate solution is generated and tested to check if it is successful. 
If it is not successful, then another candidate solution is generated to be alternative and tested until a satisfactory solution is found.

Testing the performance of the candidate solution is a crucial step in an optimization task.
Current studies generally test the performance of the candidate solution from the aspect of quality over function evaluations.
However, the computation time is also a crucial measurement, particularly the applications in the real world.
For the real optimization tasks, users generally consider more about how long the optimization algorithms cost to obtain the satisfactory solutions.
Therefore, this paper studies both quality and computation time in optimization problems.
We select and test the well-known optimization algorithms of Bayesian optimization and evolutionary algorithms on the well-known test functions of Schwefel, Griewank, Rastrigin.
These algorithms are evaluated in terms of both quality and computation time. 
The pros and cons of these algorithms are discussed from the aspect of quality and computation time.

The results indicate that Bayesian optimization performs faster convergence than other EAs in the early function evaluations.
At the end of optimization, Bayesian optimization and evolutionary algorithms perform the similar quality with enough function evaluations.
However, the results demonstrate that Bayesian optimization performs substantially increasing computation time over the number of function evaluations as that is well-known.
To solve problems faster, Bayesian optimization is suitable to the optimization problems that can be solved in the limited function evaluations.
Evolutionary algorithms perform slower convergence but the much less computation time than BO, and finally obtain the similar even higher quality at the end of optimization.
Therefore, evolutionary algorithms are suitable for the optimization problems that are allowed to be tested with enough function evaluations.

In summary, this paper analysis the optimization algorithms in terms of quality and computation time. This provides the recommendation to evaluate and select the optimization algorithms for different optimization problems by considering both quality and computation time instead of the one-for-all evaluation of quality. 
Therefore, the suitable optimization algorithms can be applied to the optimization problems, particularly the optimization tasks that are sensitive to the computation time.

\section{Related Work}
\label{sec:related}
There are many optimization algorithms that can be used to solve optimization problems. 
In general, the mainstream algorithms include Bayesian optimization and evolutionary algorithms.
These optimization algorithms are generally evaluated in terms of quality over the number of function evaluations.
Typically, Bayesian optimization is a sequential design strategy that usually employed to optimize expensive-to-evaluate functions \cite{frazier2018tutorial}.
Evolutionary algorithms are a type of heuristic optimization algorithms that are inspired by biological mechanisms of evolution, e.g., genetic algorithms, evolution strategy, Covariance matrix adaptation evolution strategy (CMA-ES), particle swarm optimization.
In this section, we review the evaluations in the existing works of optimization problems from the aspects of quality and computation time of the mainstream optimization algorithms.

The current studies of Bayesian optimization generally evaluate the performance of the optimization algorithms in terms of quality, i.e., the quality over the number of function evaluations.
Ru et al. \cite{ru2018fast} developed a new optimization algorithm of the fast information-theoretic Bayesian optimisation method to avoid the need for sampling the global minimiser, thus reducing computation time.
The performance of the optimization algorithms are evaluated on three benchmark test functions over the number of function evaluations.
Wu et al. \cite{wu2019hyperparameter} presented a hyperparameter tuning algorithm for machine learning models based on Bayesian optimization.
Similarly, the algorithm was evaluated on the well-known dataset MNIST over the number of function evaluations.
The recent works \cite{lan2021learningASC,lan2021learningNC} used Bayesian optimization to learn locomotion tasks in modular robots, which the performance is evaluated in terms of quality as well.
These Bayesian optimization algorithms are generally evaluated over the number of function evaluations for optimization problems. 
However, the computation time of Bayesian optimization grows substantially with the number of function evaluations.
Therefore, the evaluation over the number of function evaluations is unfeasible for large numbers of evaluations, which greatly limits its applicability.

In the field of evolutionary algorithms, the methods in the current studies are basically evaluated in terms of quality.
Gir{\'a}ldez et al. \cite{giraldez2005knowledge} proposed a novel data structure to reduce the time complexity of the evolutionary algorithm by a fast evaluation of examples from the dataset.
Deb et al. \cite{deb2002computationally} propose a computationally efficient evolutionary algorithm for real-parameter optimization.
The performance of these algorithms are evaluated over function evaluations. 
Thiele et al. \cite{thiele2009preference} proposed a new preference-based evolutionary algorithm for multi-objective optimization.
The quality in terms of multi-objective are used to evaluate the performance of the algorithms. 
Young et al. \cite{young2015optimizing} proposed multi-node evolutionary neural networks as a method for automating network selection.
Unsurprisingly, the performance of the method is evaluated with the quality over the number of function evaluations. 
The current studies \cite{lan2019simulated,lan2019evolutionary,lan2019evolving} applied evolutionary algorithms to learn the controllers of robots. 
However, their performance are evaluated in terms of quality. 
An recent study \cite{gao2021neat} applied the evolutionary algorithms to the multiclass classification and evaluated the performance in terms of quality.
The performance of the optimization in classification is commonly evaluated on classification accuracy. 
However, the running computation time on classification model training, parameter-tuning, and testing is missing, which is a very important concern for the real-world applications.

Interestingly, the recent study \cite{lan2021time} evaluated the efficiency of optimization algorithms with the quality over computation time.
Finally, the algorithms are evaluated from both quality and computation time.
Computation time is an important factor of the performance of the vectorization and line detection algorithms, especially in industrial applications.
Regarding the current studies of computation time in machine learning, an early study \cite{liu1998accuracy} proposed to evaluate optical flow algorithms for practical applications with a 2-dimensional scale that is a combination of quality and computation time.
Wenyin and Dori \cite{wenyin1997protocol} proposed using the elapsed time of running algorithms to evaluate the performance in terms of computation time.
Zhang et al. \cite{zhang2017up} provided an up-to-date comparison on the computation time of eleven state-of-the-art classification algorithms.
The computation time of optimization algorithms is a crucial measurement, particularly the optimization tasks in the real world.

In conclusion, the existing works of optimization problems are generally evaluated in terms of quality over the number of function evaluations, but not computation time.
We suggest that computation time should be the reasonable measurement to evaluate and select optimization algorithms.
Therefore, in this paper we investigate the quality and computation time of optimization algorithms.

\section{Methodology}
\label{sec:method}
In this paper, we select and test the well-known optimization algorithms of Bayesian optimization and evolutionary algorithms on the test functions. 
In this section, we address the well-known optimization algorithms and the test functions.

\subsection{Optimization Algorithms}
\label{subsec:algorithms}

\subsubsection{Bayesian Optimization}

\begin{algorithm}[!ht]
    \small
	\caption{Bayesian optimization}
    \label{alg:bo}
    \kwInit{$n$ initial solutions $\overrightarrow{\mathcal{X}}_1, \overrightarrow{\mathcal{X}}_2, ..., \overrightarrow{\mathcal{X}}_i, ..., \overrightarrow{\mathcal{X}}_n$;  $\overrightarrow{\mathcal{X}}_i$ is a $j$ dimensional vector $\overrightarrow{\mathcal{X}}_i(x_1, x_2, ..., x_j)$.}
    \kwEval{ to obtain fitness $f_1, f_2, ..., f_i, ..., f_n$.}
    \kwCreate{initial GP: $\mu(\overrightarrow{\mathcal{W}}_{1:n})$, $\sigma^2(\overrightarrow{\mathcal{W}}_{1:n})$.}
    \For{$k = n+1,n+2, ...$}    
    {
    \textbf{select} a new solution $\overrightarrow{\mathcal{X}}_k$ by optimizing acquisition function $u(\overrightarrow{\mathcal{X}}_k)$:
     \[ \overrightarrow{\mathcal{X}}_k = \arg\max_{\overrightarrow{X}_k}  u(\overrightarrow{\mathcal{X}}_k|\overrightarrow{X}_{1:k-1})\] \\
    \textbf{evaluate} the new solution $\overrightarrow{X}_k$ to obtain the objective value $f_k$. \\
    \textbf{augment} data $(\overrightarrow{X}_{1:k}, f_{1:k}) = \{\overrightarrow{X}_{1:(k-1)}, (\overrightarrow{X}_{k}, f_k)\}$. \\
    \textbf{update} GP: $\mu_k(\overrightarrow{X}_{1:k})$, $\sigma_k^2(\overrightarrow{X}_{1:k})$.
    }
    \Return data $(\overrightarrow{X}_{1:k},f_{1:k})$.
\end{algorithm}

Bayesian Optimization is a well-known data efficient optimization algorithm, which builds a Gaussian process (GP) approximation of the objective function $f$ of a task. 
It selects candidate solutions by the guiding of an acquisition function that balances exploration and exploitation. 
In this paper, we use \emph{Gaussian Process Upper Confidence Bound} (GP-UCB) that is a state-of-the-art acquisition function \cite{brochu2010tutorial}.
In addition, the choice of the kernel function for the GP is crucial, as it determines the shape of the functions that the GP can fit \cite{brochu2010tutorial}. 
We use Mat\'ern function kernel that is currently a popular choice.
The pseudocode of Bayesian optimization is shown in \autoref{alg:bo}.
For more details of Bayesian optimization, please refer to the literature \cite{brochu2010tutorial}.

\subsubsection{Evolutionary Algorithms}

Evolutionary Algorithms are population-based, stochastic optimization algorithms based on the Darwinian principles of evolution \cite{Eiben2015intro}. 
The optimization process works by iteratively creating a new solution. 
The main operators in the optimization process are fitness-based parent selection, reproduction through mutation and/or crossover, and fitness-based survivor selection. 
In this paper, we use the standard real-valued evolutionary algorithm, evolutionary strategy. 
For more details of evolutionary algorithms, please refer to the well-known book \cite{brochu2010tutorial}.

\begin{algorithm}[!ht]
    \small
	\caption{Typical Evolutionary Algorithms \cite{eiben2003introduction}.}
    \label{alg:hyperneat}
    \kwInit{$n$ initial solutions $\overrightarrow{X}_1, \overrightarrow{X}_2, ..., \overrightarrow{X}_i, ..., \overrightarrow{X}_n$ and the objective value $f_1, f_2, ..., f_i, ..., f_n$.}
    \For{$k = 1, 2, 3, ...$}
    {
    \uIf{$k \neq 1$}
    {
    \textbf{generate} candidate sulotions $\overrightarrow{X}_{((k-1)*n+1):(k*n)}$ by mutation and crossover.\\
    }
    \textbf{evaluate} solutions $\overrightarrow{X}_{((k-1)*n+1):(k*n)}$ to obtain the objective values $f_{((k-1)*n+1):(k*n)}$. \\
    \textbf{select} solutions from $X_{((k-1)*n+1):(k*n)}$ and their $f_{((k-1)*n+1):(k*n)}$. \\
    \textbf{update} solutions $\overrightarrow{X}_{1:(k*n)}$ and their objective values $f_{1:(k*n)}$.
    }
    \Return data ($\overrightarrow{X}_{1:(k*n)}, f_{1:(k*n)}$).
\end{algorithm}

\begin{table*}[!ht]
    \renewcommand{\arraystretch}{1.5}
    \setlength\tabcolsep{4pt}
    \begin{tabular}{l l l l l l} \toprule
        functions & definition  & Global minimum & domain & scalable & local optima \\ \midrule
        Schwefel & $f(\overrightarrow{X}) = 418.9829d - \sum_{i=1}^{d} x_i \sin{\sqrt{|x_i|}} $ & $f(\overrightarrow{X}^*) = 0, \text{at~} \overrightarrow{X}^* = (420.97,...,420.97)$ & $x_i \in [-500,500]$ & $i = 1,...,d$ & multiple \\
        Griewank & $f(\overrightarrow{X}) = \sum_{i=1}^{d} \frac{x_i^2}{4000} - \prod_{i=1}^{d} \cos{(\frac{x_i}{\sqrt{i}})} + 1$ & $f(\overrightarrow{X}^*) = 0, \text{at~} \overrightarrow{X}^* = (0,...,0)$ & $x_i \in [-600,600]$ & $i = 1,...,d$ & multiple \\ 
        Rastrigin & $f(\overrightarrow{X}) = 10d + \sum_{i=1}^{d} [x_i^2 - 10cos(2\pi x_i)]$ & $f(\overrightarrow{X}^*) = 0, \text{at~} \overrightarrow{X}^* = (0,...,0)$ & $x_i \in [-5.12,5.12]$ & $i = 1,...,d$ & multiple \\ \bottomrule
    \end{tabular}
    \caption{The objective functions and their characteristics.}
    \label{tab:functions}
\end{table*}

\subsection{Test Functions}
\label{subsec:functions}

A optimization problem is about finding an optimal solution $\overrightarrow{X} = (x_1,...,x_n) \in \mathcal{X}$ of an objective function $f: \mathcal{X} \rightarrow \mathbb{R}$, that is:
\begin{equation}
    \overrightarrow{X}^{*} = \arg \max_{\overrightarrow{X} \in \mathcal{X}} f(\overrightarrow{X}) ,
\end{equation}
where the analytical form of the objective function $f(\overrightarrow{X})$ is usually (but not necessarily) unknown, i.e., black-box optimization, in the real optimization tasks \cite{lan2021time,bossek2017smoof}.
To clearly observe and analysis the quality and computation time of the various dimensional optimization, we select many benchmark test functions as the objective functions.
The benchmark test functions should meet the requirements of:
\begin{itemize}
    \item scalable, the dimension of a solution $\overrightarrow{X} = (x_1,...x_n)$ can be various.
    \item given global minimum, which can be observed clearly.
    \item multimodal and multiple local optima, which are sufficiently challenging so that these algorithms cannot quickly converge to the optimal solution.
\end{itemize}

We select these test functions that are from the well-known Black-Box Optimization Benchmark (BBOB) \cite{hansen2009real}.
The source code of these objective functions can be found in \footnote{\url{https://www.sfu.ca/~ssurjano/optimization.html}}.
In this paper, we finally selected the function Schwefel, Griewank, and Rastrigin.
Specifically, we set the dimension of the test functions as ten, i.e., $n = 10$ in the solution $\overrightarrow{X} = (x_1,...x_n)$.
The selected test functions and their characteristics are shown in \autoref{tab:functions}.

\section{Results and Discussion}
\label{sec:results}

In this section, we present the quality and the computation time of the well-known optimization algorithms on the test functions.

\begin{figure*}[!ht]
    \centering
    \large
    \begin{adjustbox}{max width=0.99\textwidth}
    \begin{tabular}{c c c}
        \includegraphics[width=0.4\textwidth]{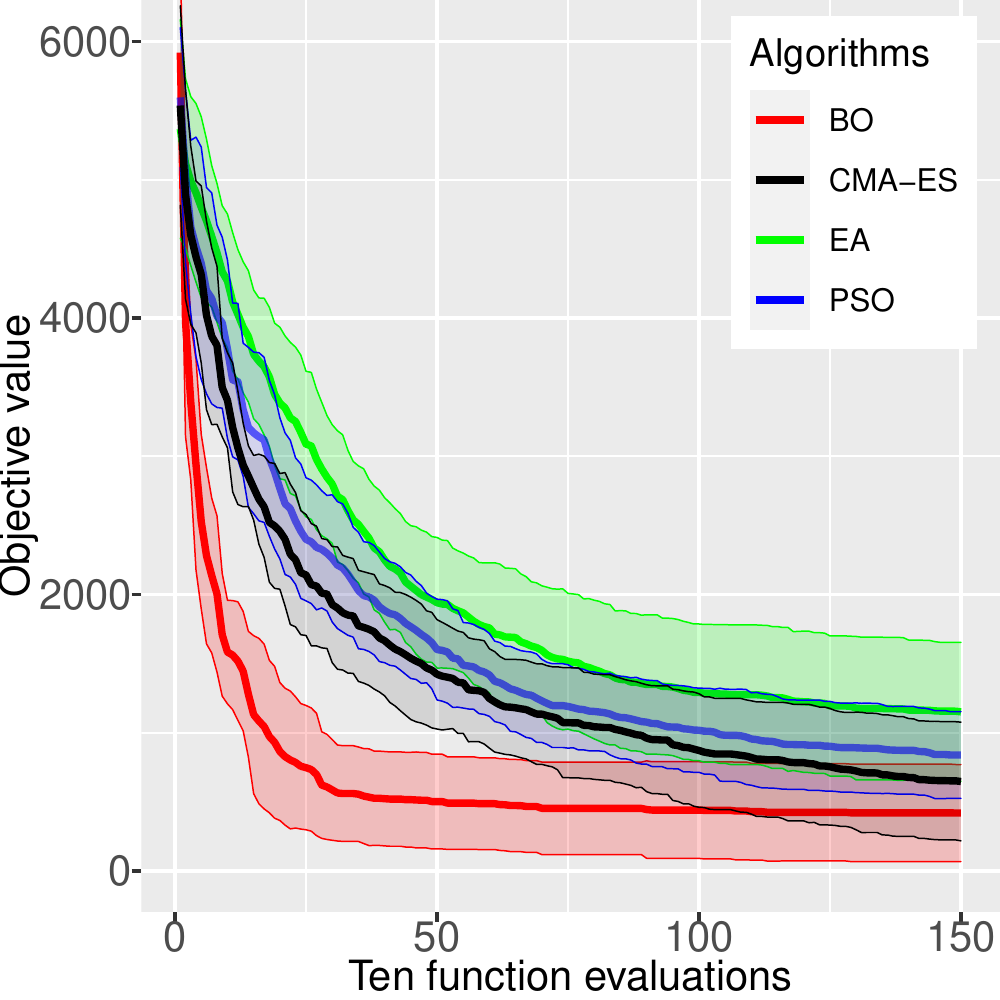} &
        \includegraphics[width=0.4\textwidth]{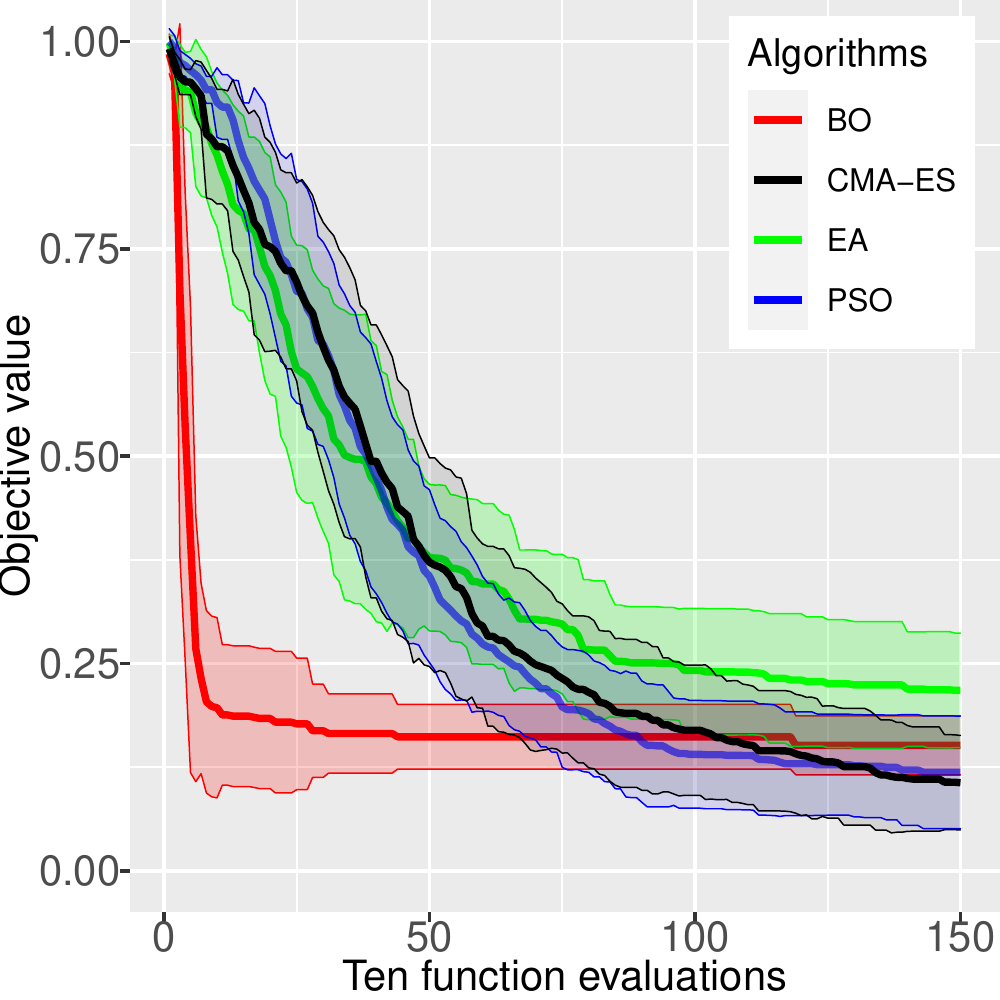} &
        \includegraphics[width=0.4\textwidth]{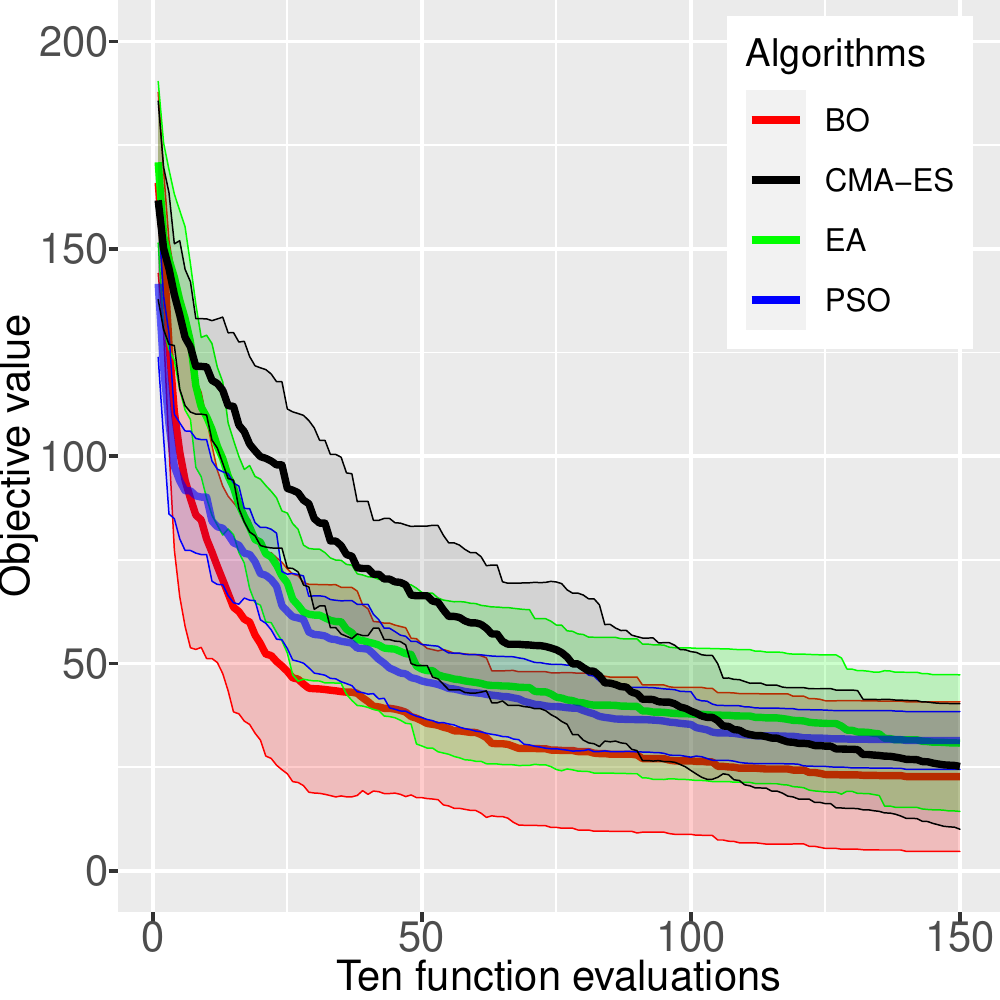} \\
        Schwefel & Griewank & Rastrigin \\
    \end{tabular}
    \end{adjustbox}
    \caption{The optimization performance of the Bayesian optimization and evolutionary strategy in terms of quality over ten function evaluations (one generation in the EAs) averaged over ten runs on the function Schwefel, Griewank, and Rastrigin. The solid lines represent the average objective value in ten runs. The shadow intervals represent the confidence of the average objective value.}
    \label{fig:quality}
\end{figure*}

\subsection{Results in Quality}

We run the optimization algorithms of BO, CMA-ES, EA (ES), and PSO on the ten dimensions of test functions, Schwefel, Griewank, Rastrigin. 
In optimization problems, quality is a mainstream metric to evaluate the performance of the optimization algorithms. 
We run these optimization algorithms ten times on the test functions and take the average of the objective value over ten function evaluations (one generation in the EAs), as shown in \autoref{fig:quality}.
The solid lines represent the average objective values that convergence to the global minimum of these test functions.
The shadow intervals represent the confidence of the average objective value over the ten runs.
In the early stage, BO performs better quality than other EAs during the given function evaluations, i.e., faster convergence.
However, BO 
Although the EAs (CMA-ES, EA (ES), and PSO) search the global minimum slower than BO, they convergence gradually during the process and finally obtain the similar quality as BO. 
Therefore, we conclude that BO is suitable to be applied in the optimization tasks that need to be solved in the limited function evaluations.
For the optimization tasks that are allowed to be solved in a lot of function evaluations, both BO and the EAs obtain the similar performance in terms of quality.
Importantly, the applications of optimization tasks in the real world have to consider the computation time that we analysis in the next section.

\subsection{Results in Computation Time}

\begin{figure*}[!ht]
    \centering
    \large
    \begin{adjustbox}{max width=0.99\textwidth}
    \begin{tabular}{c c c}
        \includegraphics[width=0.4\textwidth]{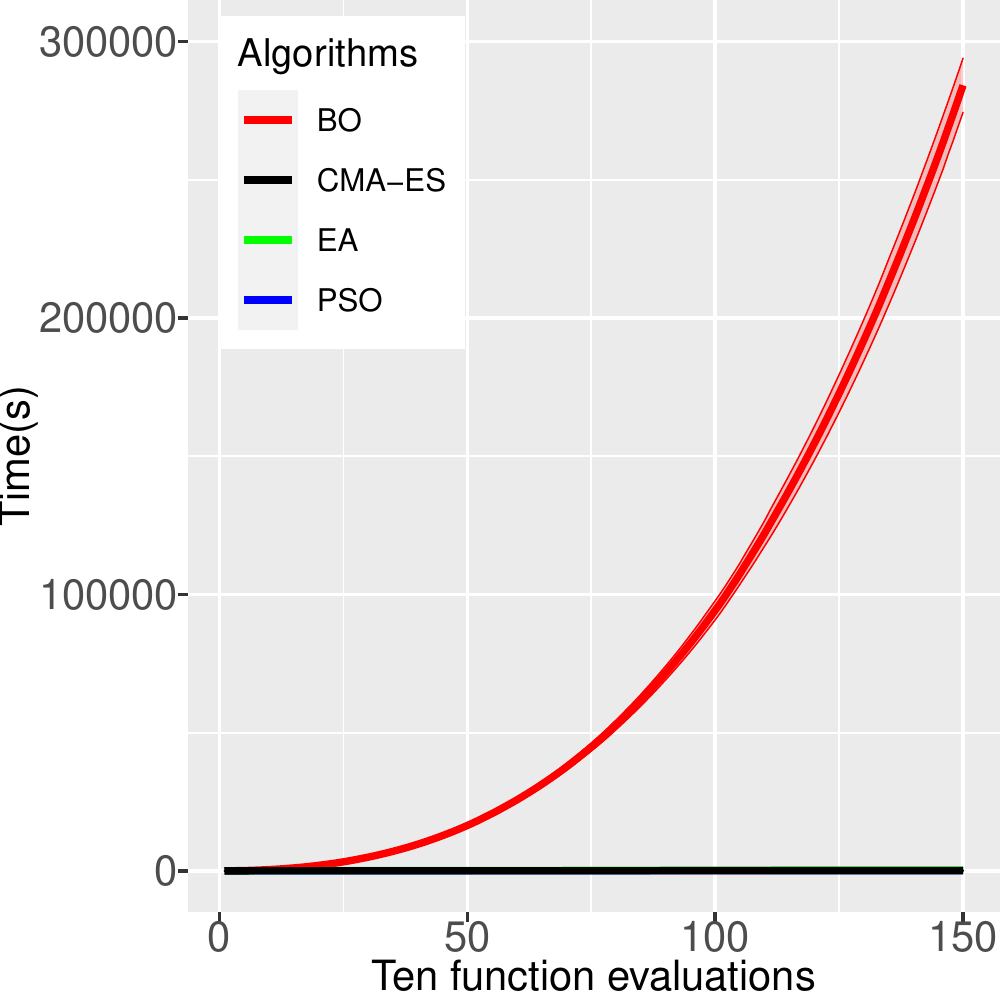} &
        \includegraphics[width=0.4\textwidth]{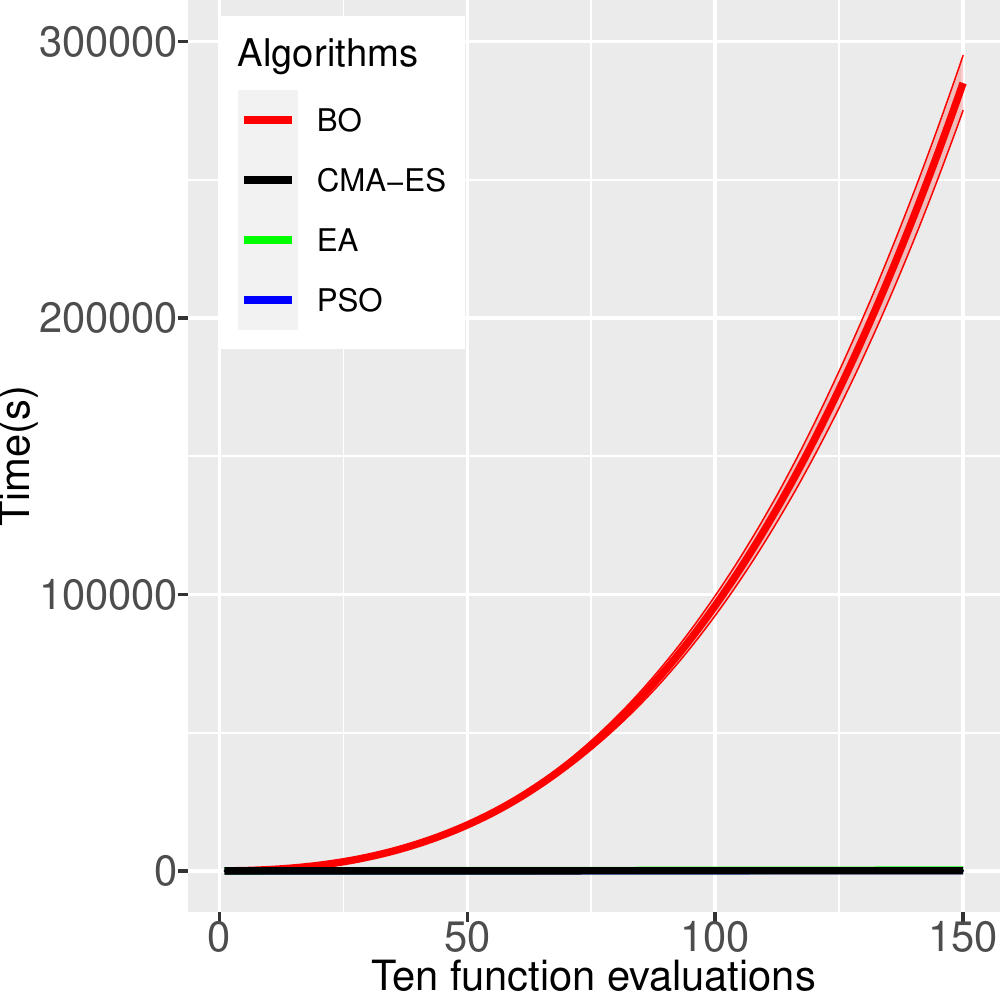} &
        \includegraphics[width=0.4\textwidth]{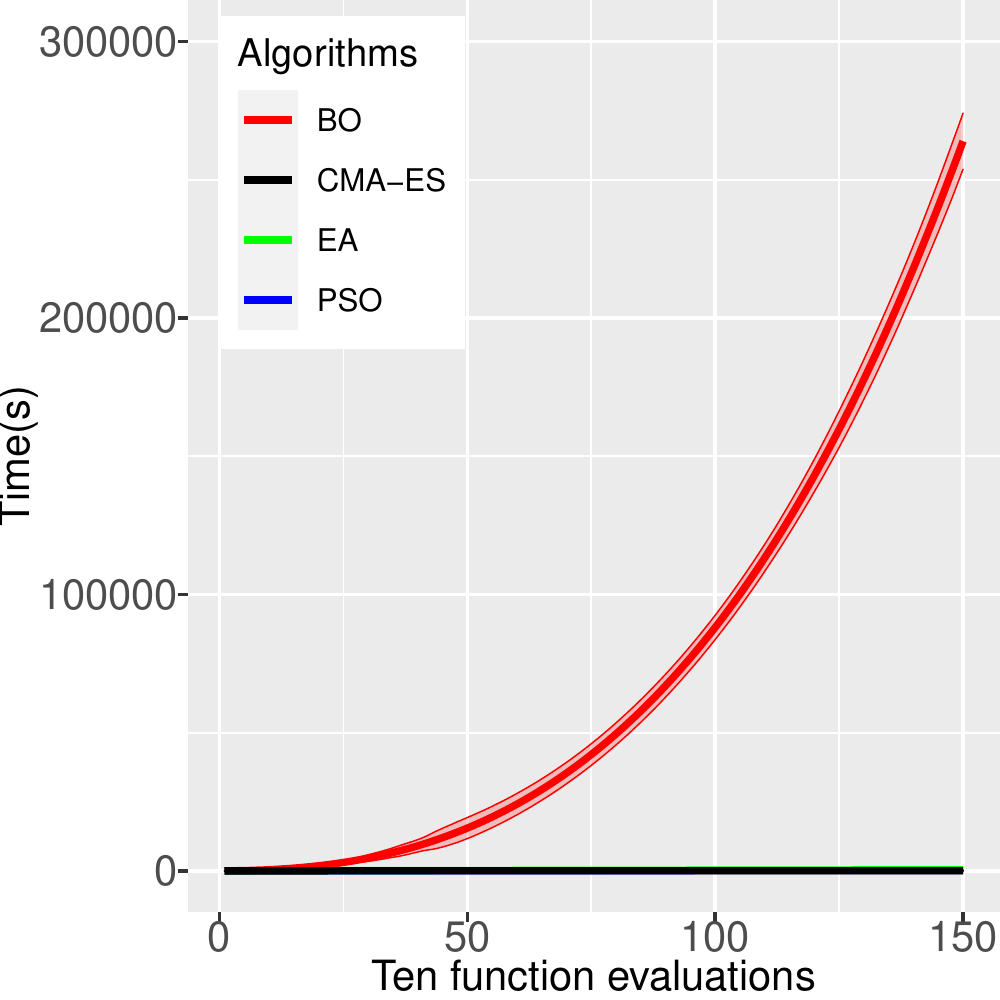} \\
        Schwefel & Griewank & Rastrigin \\
    \end{tabular}
    \end{adjustbox}
    \caption{The computation time of the optimization algorithms over ten function evaluations (one generation in the EAs) averaged on the function Schwefel, Griewank, and Rastrigin. The solid lines represent the average computation time over ten runs. The shadow intervals represent the confidence of the average computation time. Note that the EAs (CMA-ES, EA, PSO) perform the similar computation time cost, which their lines overlap in the black lines.}
    \label{fig:ct}
\end{figure*}

\begin{figure*}[!ht]
    \centering
    \large
    \begin{adjustbox}{max width=0.99\textwidth}
    \begin{tabular}{c c c}
        \includegraphics[width=0.4\textwidth]{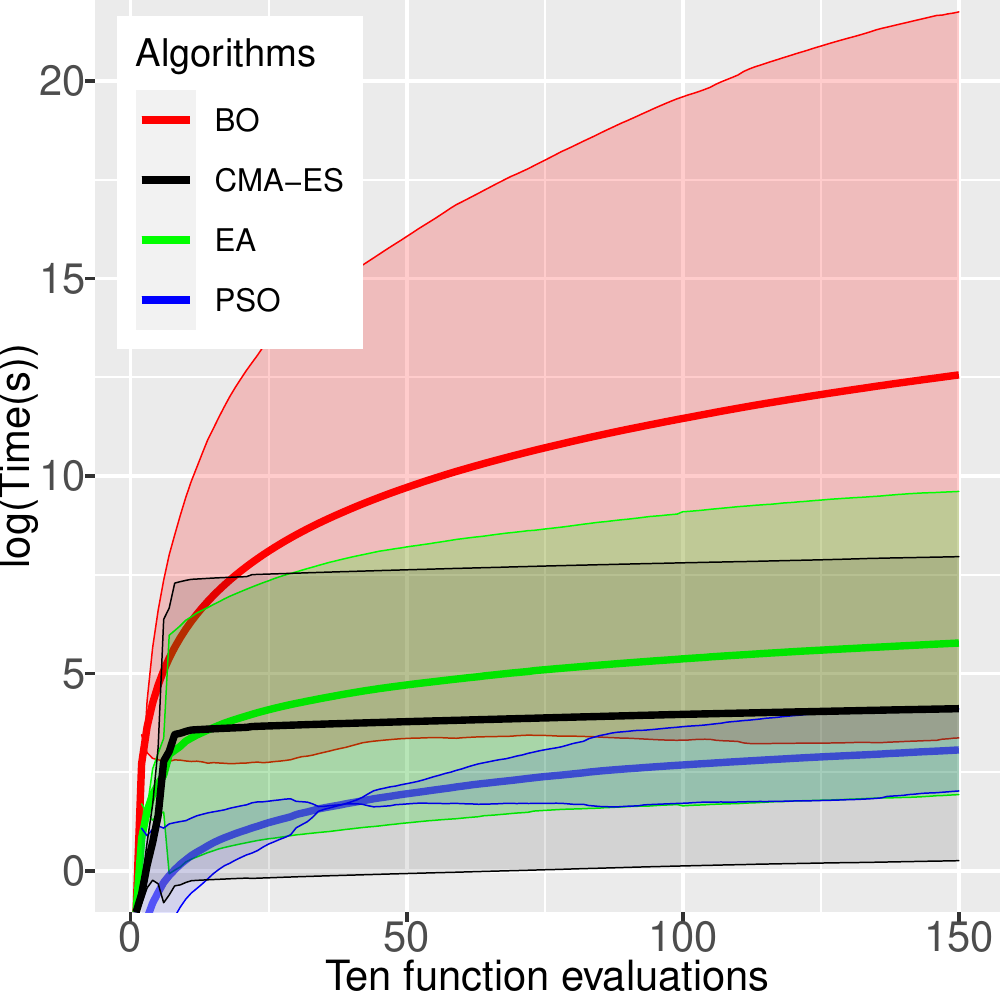} &
        \includegraphics[width=0.4\textwidth]{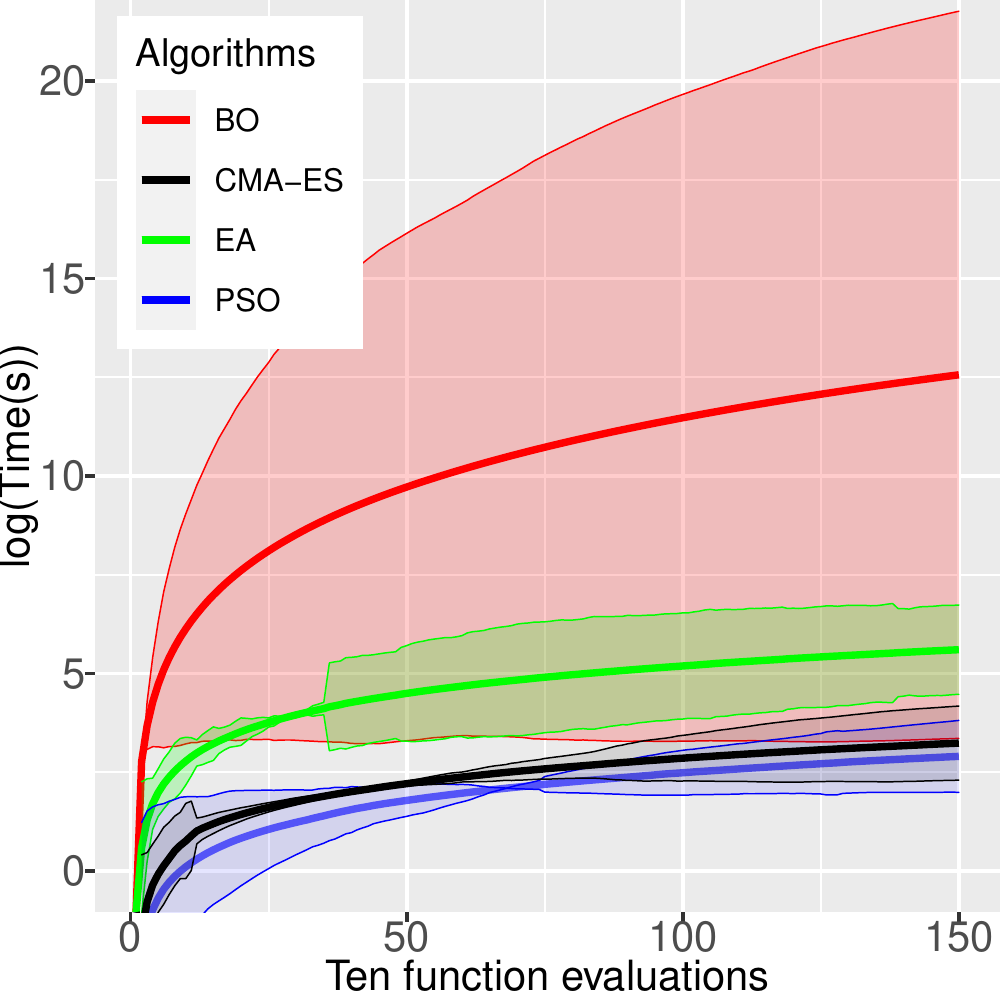} &
        \includegraphics[width=0.4\textwidth]{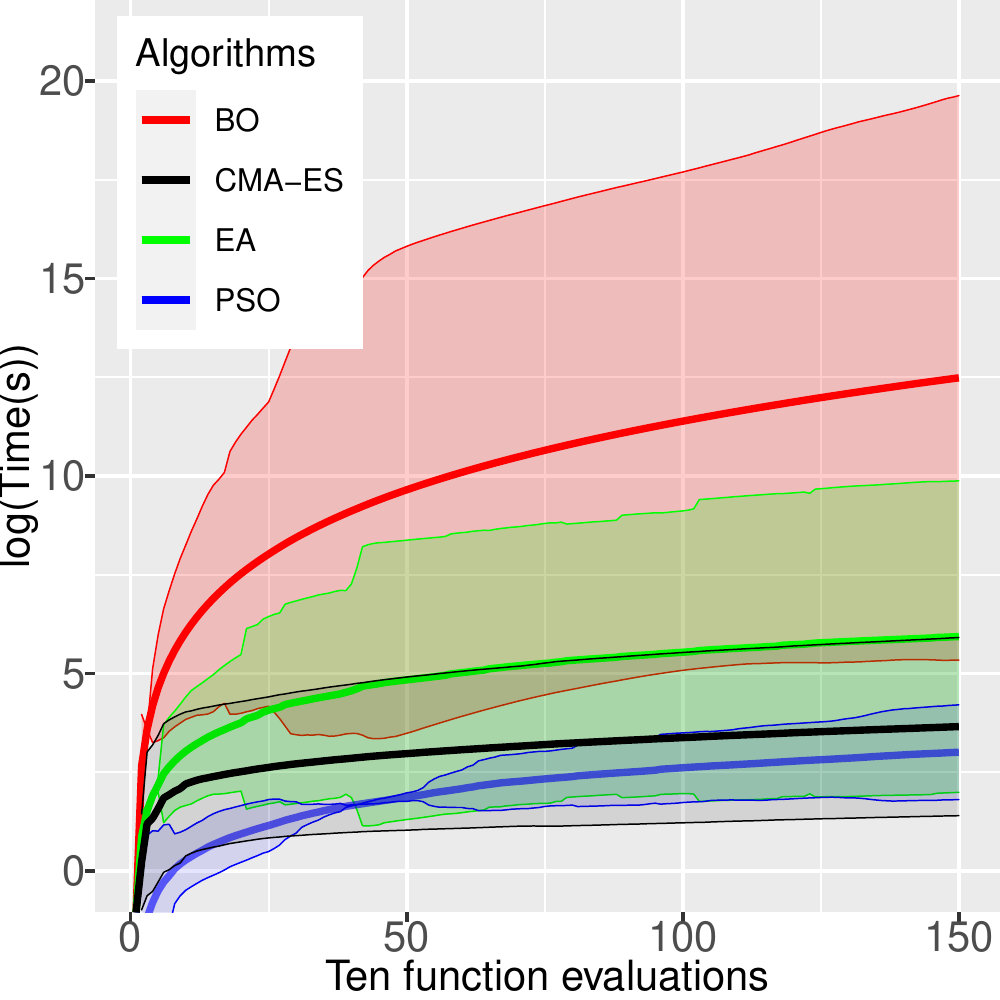} \\
        Schwefel & Griewank & Rastrigin \\
    \end{tabular}
    \end{adjustbox}
    \caption{The details in the logarithmic of computation time of the optimization algorithms over function evaluations averaged on the function Schwefel, Griewank, and Rastrigin. The solid lines represent the averaged logarithmic of computation time over ten runs. The shadow intervals represent the confidence of the averaged logarithmic of computation time.}
    \label{fig:ct_log}
\end{figure*}

In the applications of optimization tasks in the real world, users actually consider about how long the optimization algorithms obtain a desired solution.
Therefore, the computation time is one of the crucial measurements to evaluate performance of optimization algorithms.
It is well-known that the computation time of BO grows substantially with the number of evaluations, limiting its feasibility for longer optimization runs \cite{lan2021time}.
In this section, we analysis the computation time of BO, CMA-ES, EA, and PSO on the test functions. 
We collect the time stamp and thus calculate the computation time of each ten function evaluations.
The computation time of these optimization algorithms are presented in \autoref{fig:ct}. 
The solid lines represent the average computation time over ten runs, and the shadow intervals represent the confidence of the average computation time.
It is clear that BO performs substantially increasing computation time over the number of function evaluations.
In the end of the optimization, BO costs significant more than the other algorithms. 
The observed results show that BO finally cost about 280000 seconds (77 hours) during the optimization because the substantially increasing computation time over the number of function evaluations, which is also demonstrated in \autoref{fig:ct}.
These EAs cost little of computation time during the optimization.
For the observed results, these EAs cost about from 20 to 200 seconds during the optimization of 1500 function evaluations.
The lines of the computation time of the EAs (CMA-ES, EA, PSO) overlap each other.
Therefore, the computation time of these EAs are unable to be observed in \autoref{fig:ct}.
To solve this issue, we apply the logarithmic to analysis the computation time of these EAs.
The $log(Time(s))$ over ten function evaluations are shown in \autoref{fig:ct_log}.
It is clear the logarithmic of the computation time of these EAs are distinguishable. 
The EA generally cost more computation time than CMA-ES and PSO, specifically cost about 200 seconds during the optimization.
CMA-ES and PSO finish the optimization with the computation time of about 20 seconds.

In conclusion, although the different optimization algorithms obtain the optimal solutions towards the global minimum with the enough function evaluation, they perform significant different computation time.
BO performs the significant increasing computation time, while the EAs convergence to the optimal in the short computation time of 20 $\sim$ 200 seconds.
For the applications of optimization tasks in the real world, BO is suitable to be applied in the tasks that are needed to obtained desired quality in a limited function evaluations, and the EAs are suitable to search the optimal of the tasks that are allowed to be run with enough function evaluations.

\section{Conclusion}
\label{sec:conclusion}

In this paper, we proposed that the performance of optimization algorithms should be evaluated from the aspects of quality and computation time. 
The computation time is a crucial measurement for the optimization tasks in the real world applications.
We select and run the well-known optimization algorithms of Bayesian optimization and evolutionary algorithms (including CMA-ES, EA, and PSO) on the well-known test functions of Schwefel, Griewank, and Rastrigin.
The performance of these algorithms are analysed from the aspects of quality and computation time.
The results indicated that Bayesian optimization is suitable for the optimization tasks that are needed to obtained desired quality in a limited function evaluations, and the EAs are suitable to search the optimal of the tasks that are allowed to be run with enough function evaluations.
This work provides the evaluation and the selection of the optimization algorithms for different optimization problems.
In the future, we will combine the quality and computation time into a measurement that can be used to evaluate the optimization algorithms comprehensively.
In addition, we will consider the results of quality and computation time for the optimization in artificial intelligent applications of robotics \cite{lan2019evolutionary,lan2019simulated},
intelligent sensing \cite{lan2015bayesian,lan2017development},
virtual reality \cite{lan2016developmentVR,lan2016developmentUAV_VR},
autonomous driving \cite{xu2019online} to improve the performance of these studies.


\bibliographystyle{IEEEtran}
\bibliography{bibliography}

\end{document}